\documentclass[runningheads]{llncs}
\usepackage[T1]{fontenc}
\usepackage{graphicx}
\usepackage{amsmath}
\usepackage{amssymb}
\usepackage{multirow}
\usepackage{graphicx}

\usepackage{algorithm}
\usepackage{algorithmic}
\usepackage{multirow}
\usepackage{graphicx}

\usepackage{amsmath}
\usepackage{multirow}
\usepackage{graphicx}
\usepackage{stix,bbding,pifont,utfsym,fontawesome}
\usepackage{cuted}
\usepackage{amsmath}
\usepackage{amsfonts}
\usepackage{listings}
\usepackage{xcolor}

\definecolor{keyword}{rgb}{0.58,0,0.82}
\definecolor{comment}{rgb}{0.5,0.5,0.5}
\definecolor{string}{rgb}{0.56,0.93,0.56}

\lstdefinestyle{mystyle}{
    backgroundcolor=\color{white},
    commentstyle=\color{comment},
    keywordstyle=\color{keyword},
    numberstyle=\tiny\color{gray},
    stringstyle=\color{string},
    basicstyle=\ttfamily\footnotesize,
    breaklines=true,
    captionpos=b,
    numbers=left,
    numbersep=5pt,
}

\begin{document}

\title{GASCADE: Grouped Summarization of Adverse Drug Event for Enhanced Cancer Pharmacovigilance}
\author{Sofia Jamil\inst{1} \and 
Aryan Dabad\inst{1} \and 
Bollampalli Areen Reddy\inst{1} \and 
Sriparna Saha\inst{1} \and 
Rajiv Misra\inst{1} \and 
Adil A. Shakur\inst{2}}
\authorrunning{S. Jamil et al.}
%
\institute{Department of Computer Science \& Engineering, Indian Institute of Technology Patna, India\\
\email{\{sofia\_2321cs16, aryan\_2101ai36, bollampalli\_2101cs19, sriparna, rajivm\}@iitp.ac.in}
\and
Indira Gandhi Institute of Medical Sciences, Patna\\
\email{\{adilshakur04\}@gmail.com}}
\maketitle              
\begin{abstract}
In the realm of cancer treatment, summarizing adverse drug events (ADEs) reported by patients using prescribed drugs is crucial for enhancing pharmacovigilance practices and improving drug-related decision-making. While the volume and complexity of pharmacovigilance data have increased, existing research in this field has predominantly focused on general diseases rather than specifically addressing cancer. This work introduces the task of grouped summarization of adverse drug events reported by multiple patients using the same drug for cancer treatment. To address the challenge of limited resources in cancer pharmacovigilance, we present the \textit{\textbf{MultiLabeled Cancer Adverse Drug Reaction and Summarization (MCADRS)}} dataset. This dataset includes pharmacovigilance posts detailing patient concerns regarding drug efficacy and adverse effects, along with extracted labels for drug names, adverse drug events, severity, and adversity of reactions, as well as summaries of ADEs for each drug. Additionally, we propose the \textit{\textbf{Grouping and Abstractive Summarization of Cancer Adverse Drug events (GASCADE)}} framework, a novel pipeline that combines the information extraction capabilities of Large Language Models (LLMs) with the summarization power of the encoder-decoder T5 model. Our work is the first to apply alignment techniques, including advanced algorithms like Direct Preference Optimization, to encoder-decoder models using synthetic datasets for summarization tasks. Through extensive experiments, we demonstrate the superior performance of \textit{GASCADE} across various metrics, validated through both automated assessments and human evaluations. This multitasking approach enhances drug-related decision-making and fosters a deeper understanding of patient concerns, paving the way for advancements in personalized and responsive cancer care. \textit{The code and dataset used in this
work are publicly available.\footnote{\url{https://github.com/SofeeyaJ/GASCADE\_ECIR2025}}}.


\keywords{Pharmacovigilance  \and Adverse Drug Reaction \and Summarisation.}
\end{abstract}

\section{Introduction}

Cancer has rapidly escalated to become the world’s second-leading cause of death, imposing an immense and growing burden on global health. In 2023 alone, the National Center for Health Statistics reported 1,958,310 new cancer cases and 609,820 cancer-related deaths in the United States, highlighting the urgent need for effective treatments. The efficacy of drugs is a crucial aspect of cancer treatment, but the adverse drug events that accompany their use pose significant challenges in determining their true effectiveness.
This set of activities, focused on detecting, evaluating, understanding, and preventing any adverse effects or additional concerns associated with drugs, is known as \textbf{Pharmacovigilance}. A key challenge in cancer pharmacovigilance is the precise collection of information on the drugs patients use, the adverse drug reactions (ADRs) \footnote{In this paper, Adverse Drug Reactions (ADRs) and Adverse Drug Events (ADEs) are used interchangeably.} they experience, and the severity of those reactions. Recently, the adoption of ADR monitoring systems has grown significantly \cite{singh2017adverse,shareef2015development,hou2016national}, signaling an increased focus on managing ADRs. With more individuals sharing their personal medication experiences on social media, mining these texts for ADR information has become increasingly valuable \cite{karimi2015text,sarker2015portable}. 
With ADE accounting for 1.3 million visits to the emergency department in the United States alone \cite{cdc}, a key application of collaboration can be the information pertaining to the ADRs of the medications prescribed for treating cancer so that the doctors are more aware of what is going to come its way. 
In this context, an ADR summarization tool could efficiently extract and condense user-reported experiences, offering insights into the adverse reactions associated with each cancer drug. 
\begin{figure}[!htbp]
\vspace{-0.3cm}
\centerline{\includegraphics[scale=0.05]{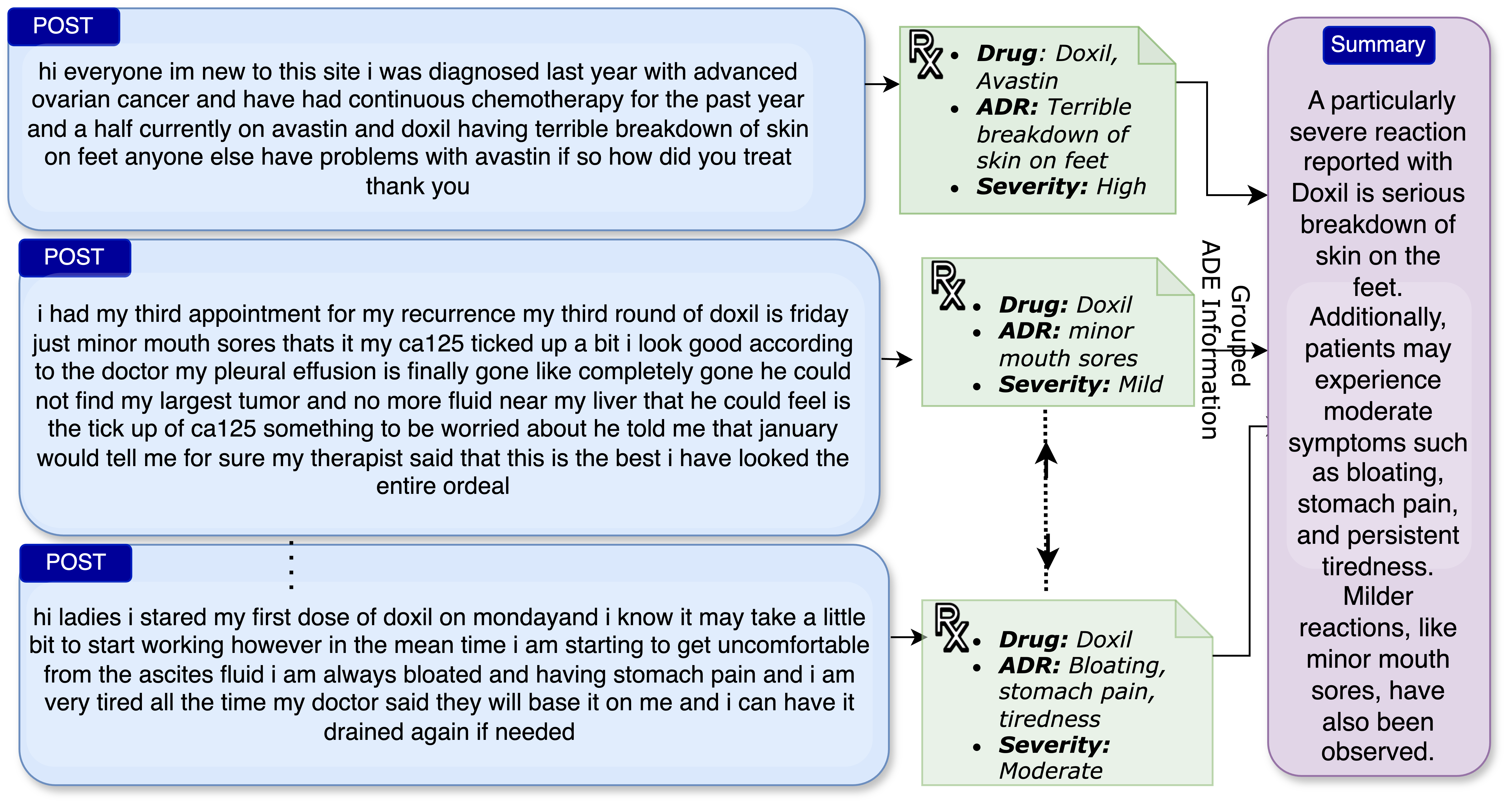}}
\caption{ Our \textit{GASCADE} Framework summarising the Adverse Drug Event Information for each drug}
\label{motivation}
\vspace{-0.7cm}
\end{figure}

This motivated our study on the task of summarizing ADE related to cancer medications. We propose a novel two-phase approach for extracting and summarizing ADRs, introducing a dataset curated from online cancer forums where patients discuss drugs used in cancer therapy and their experiences. Our work represents the first attempt to integrate adverse drug event information extraction with summarization, particularly for cancer-related drugs. By emphasizing ADR summarization, healthcare providers can better anticipate potential side effects and make more informed medical decisions. 
Furthermore, understanding the causes and nature of ADRs enables drug manufacturers to refine drug formulations, potentially improving their efficacy. 
This work aims to enhance access to high-quality healthcare and support health equity, aligning with the \textbf{Sustainable Development Goals (SDG)} proposed by UNESCO.


Social media texts are inherently noisy, using colloquial language, slang, metaphors, and non-standard grammar, which complicates biomedical data extraction. Despite advancements, NLP models still struggle with the rapid evolution of language and distinguishing indications from ADEs. Deep learning approaches have improved ADE and entity extraction, \cite{liu2019towards}, but challenges remain, such as distinguishing between indications (reasons for taking a drug) and ADEs (unintended side effects). The advent of Large Language Models (LLMs), trained on extensive datasets across various writing styles and medical content, offers new opportunities in the clinical domain, including medical summarization \cite{ghosh_summarisation}, question summarisation \cite{ghosh_question}, and summarizing code-mixed queries \cite{ghosh_code_mix}.




However, extracting critical medical information for cancer pharmacovigilance has received limited attention, largely due to the scarcity of cancer-specific data available for LLM pretraining. To address this gap, we introduce the \textit{MultiLabeled Cancer Adverse Drug Reaction and Summarization (MCADRS)} dataset, which consists of 2,000 posts discussing cancer drugs and their associated adverse drug events, complete with adversity and severity labels as well as corresponding ADE summaries for 791 unique drugs. We benchmarked this dataset using our proposed framework, \textit{Grouping, and Abstractive Summarization of Cancer Adverse Drug Events (GASCADE)}. It is a comprehensive two-step framework that generates textual embeddings through large language models and employs QLoRA \cite{Qlora}, a low-rank adapter technique for efficient fine-tuning and applies a precise inference process to extract drug names, identify ADEs, and assess the severity and adversity of these reactions. The extracted ADEs are then grouped by drug, with sub-grouping based on the severity of reactions, ranging from high to mild. The summarization module uses advanced post-processing alignment techniques, including Direct Preference Optimization (DPO) \cite{DPO}, with synthetically generated preference datasets specifically designed for ADE summarization. A pipeline illustrating the \textit{GASCADE} framework is shown in Figure \ref{motivation}. \textbf{Our contributions} include:

 \textbf{a.} A novel task of ADE summarization for specific cancer drugs aimed at generating summaries of ADE ranked by severity, from high to mild.

 \textbf{b.} The introduction of the \textit{MCADRS} dataset, which supports further research in cancer pharmacovigilance.

 \textbf{c.} The development of \textit{(GASCADE)}, an advanced framework that integrates pre-trained language models, efficient fine-tuning methodologies like QLoRA, and alignment algorithms such as DPO for improved ADE summarization.

 \textbf{d.} Comprehensive human evaluations supported by in-depth qualitative analyses and risk assessments. This rigorous approach ensures the safety, reliability, and effectiveness of our model, affirming its readiness for real-world healthcare applications with a high degree of confidence.

\section{Related Works}
ADE extraction and classification have long been a focus of pharmacovigilance research. Early works \cite{rl19,rl20,rl21} introduced machine learning techniques using feature engineering and word embeddings for detecting ADEs in social media texts.
Deep learning models for ADR detection were investigated by Mohammadi et al. \cite{rl24}, using data from PubMed and Drug Central, showing that Transformer-based models surpassed traditional approaches in ADR detection. 
Similarly, Li et al. \cite{rl26} proposed an adversarial transfer learning method for identifying ADRs and Hussain et al. \cite{rl25} developed a system for ADR detection in clinical texts by fine-tuning BERT using the Framework for Adapting Representation Models (FARM). 
Following the success of BERT, various architectures were introduced into biomedical NLP, proving effective for ADE detection tasks, such as SpanBERT \cite{rl22}. 

\textbf{Limitations:} Most research on ADEs has centered on generic conditions, often overlooking the unique aspects of ADRs associated with specific diseases. Additionally, prior studies have neglected assessing ADR severity, crucial for targeted care. Our work addresses these gaps by developing a cancer-specific pharmacovigilance model that incorporates ADR severity to enhance patient care and management.

\section{MultiLabeled Cancer Adverse Drug Reaction and Summarization (MCADRS) Dataset}
Our research began with a comprehensive review of existing literature to identify datasets related to adverse drug events (ADEs). Prior to this work, there were no datasets focused on summarizing cancer pharmacovigilance information that detailed ADEs linked to various cancer drugs. We address this gap by presenting a new unimodal dataset featuring 791 cancer drug names. Table \ref{comparison} compares our proposed dataset with other ADE-related datasets. The dataset creation and validation were guided by medical doctors, who also co-authored this paper. The steps to construct this corpus are outlined as follows: 
\newline \textbf{Data Collection: } We conducted a recent qualitative analysis of online health discussion forums to capture cancer patients' experiences with adverse drug reactions (ADR) from chemotherapy treatments. After an extensive review, we selected two key resources: Cancer Research UK (CRU) \footnote{\url{https://www.cancerresearchuk.org/}} and Cancer Survival Network (CSN) \footnote{\url{https://csn.cancer.org/}}. These forums provided access to content without requiring membership and directly promoted discussion among patients about side effects and experiences with drug consumption or discontinuation. 
We performed internal searches on these forums using keywords like “side effects,” “adverse drug reactions,” “adverse drug events,” and “drug reactions.” Python’s Selenium library facilitated web scraping for pharmacovigilance purposes. This process yielded approximately 1,100 results from CSN and 1,900 from CRU. The extracted data included discussion topics, cancer types, patient posts about drugs and their effects, and timestamps. To protect user privacy, we anonymized all usernames. 
\newline \textbf{Data Annotation: } 
We randomly selected 100 samples from the dataset and provided them to medical experts, who developed annotation guidelines for two main tasks: converting data into labeled format and writing gold-standard summaries for each drug. We recruited two medical students and one Ph.D. student, selected based on criteria such as age (25+), fluency in English, and willingness to handle sensitive material. The annotation process was completed in two months, and participants were compensated for their efforts. \footnote{ The medical students were remunerated with gift vouchers and honorariums in accordance with \url{https://www.minimum-wage.org/ international/india.}}. To label the quality of the annotated data, medical experts established standards that each sample had to meet:
\textbf{- }For each post mentioning multiple drugs and numerous effects (positive and negative), extract only those drug names linked to adverse drug events (negative effects).

\textbf{- } Each data instance's adversity of the drug event is assessed using specific terms indicating adversity, such as "bad," "worse," "unbearable," "irrecoverable," "permanent," or similar expressions conveying similar sentiments.
  
\textbf{- }Each data instance's severity of the drug event is assessed based on explicit mentions of congenital anomalies, life-threatening situations, disabilities, or hospitalizations (initial or prolonged). If these criteria are not explicitly stated, the severity is categorized as not applicable to that specific data point.
\newline To maintain consistency among annotators, final labels were assigned via majority voting. Annotators were instructed to remain objective without bias related to demographics or other factors. The quality of annotations was determined by calculating inter-annotator agreement (IAA) using Cohen's Kappa \cite{kappa}. The agreement score of 0.75 confirms that the annotations are acceptable and of high quality. 
\begin{table}[!htbp]
\centering
\resizebox{\columnwidth}{!}{%
\begin{tabular}{lcccccc}
\hline
\textit{\textbf{Dataset Name}} & \textbf{\begin{tabular}[c]{@{}c@{}}Corpus Size\end{tabular}} & \textbf{Language} & \textbf{\begin{tabular}[c]{@{}c@{}}Disease
Specific\end{tabular}} & \textbf{Severity} & \textbf{\begin{tabular}[c]{@{}c@{}}ADR Specific\end{tabular}} & \textbf{Text Type} \\ \hline
Gurulingappa et al. \cite{gurulingappa} & 2972 & English & \XSolidBrush  & \XSolidBrush  & \checkmark & MEDLINE case reports \\
Oronoz et al. \cite{oronoz} & 75 & Spanish & \XSolidBrush  & \XSolidBrush  & \XSolidBrush  & Clinical Discharge Reports \\
Patki et al. \cite{patki} & 10,822 & English & \XSolidBrush  & \XSolidBrush  & \checkmark & Social Media Tweets \\
Li et al. \cite{li} & 1500 & English & \XSolidBrush  & \XSolidBrush  & \XSolidBrush  & PubMed Articles \\
\textit{MCADRS}(Ours) & 2000 & English & \checkmark & \checkmark & \checkmark & Healthcare Forums \\ \hline
\end{tabular}%
}
\vspace{-0.1cm}
\caption{Comparison of Existing Datasets for Pharmacovigilance}
\label{comparison}
\end{table}
\vspace{-1cm}
\newline \textbf{Writing Gold Summaries: }
Following the data labeling, annotators crafted summaries for each drug name in accordance with medical team's  guidelines. Summaries were organized alphabetically, with the most severe ADEs listed first, followed by moderate and then mild cases. This phase required an additional month to complete. To ensure quality, our team of medical experts reviewed 10\% of the summaries, assigning ratings from 1 to 5. The summaries with the highest average ratings were selected as the final gold-standard versions.

\section{Problem Statement}

\begin{figure}[!htbp]
\centerline{\includegraphics[scale=0.05]{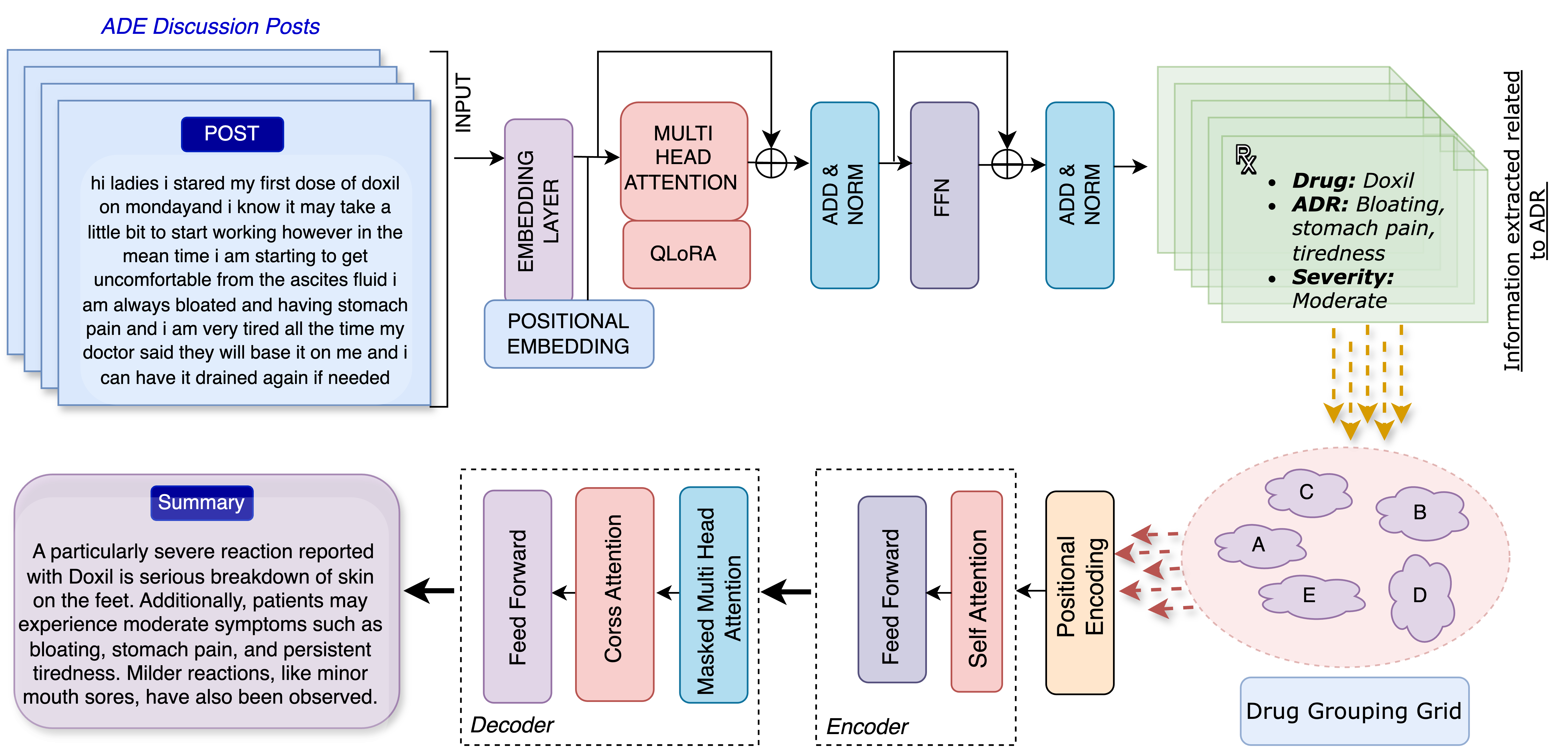}}
\caption{Proposed Model Architecture of \textit{Grouping and Abstractive Summarization of Cancer Adverse Drug events (GASCADE)}.}
\label{gascade}
\vspace{-0.7cm}
\end{figure}

Given a set of \(n\) posts discussing Adverse Drug Reactions (ADRs) and drug names, represented as,
\(
P = \{ p_1, p_2, p_3, \ldots, p_n \},
\)
where each post \( p_i \) varies in length. The process starts with extracting medical information (\( I_1, I_2, \ldots, I_k \)) from the \( n \) posts, followed by grouping this information into \( k \) clusters based on the drugs mentioned. This results in,
\(
G = \{ G_1, G_2, \ldots, G_k \},
\)
where each \( G_j \) is a cluster that contains pharmacovigilance related information about a specific drug \( j \), and collectively, these clusters cover the entire dataset (\( G_1 \cup G_2 \cup \ldots \cup G_k = D \)). For each drug cluster \( G_j \), abstractive summarization is performed, generating a summaries \(
S = \{ S_1, S_2, \ldots, S_k \},
\)
with each \( S_j \) summarizing the ADRs for drug \( j \). We introduce the \textit{\textbf{Grouping and Abstractive Summarization of Cancer Adverse Drug Events (GASCADE)}} framework to address the challenge of ADE summarization. Our approach is divided into four key steps for better understanding: (i) ADE Extraction Module, (ii) Drug Grouping Module, (iii) ADE Summarization Unit, and (iv) Alignment using DPO. Figure \ref{gascade} provides an architectural representation of our proposed \textit{GASCADE} architecture.

\noindent{\textbf{I. ADE Extraction Module: }}
To extract ADE related information, we designed an information extraction pipeline. We formulated the task as follows: Given a dataset of medical instances 
\(\{(P_i, D_i, E_i)\}_{i=1}^N\), where \(P_i\) represents the user post, 
\(D_i\) denotes the associated drug name(s), \(E_i\) is the adverse drug event (ADE), and \(S_i\) signifies the severity level, our 
goal is to develop a generative framework \(M\). This model is designed to extract the triplet \((E_i, D_i, S_i)\) ADE, drug names, and severity level based on a given user post \(P_i\). 
\begin{equation}
    M: P_i \rightarrow (E_i, D_i, S_i) 
\end{equation}
where \(M\) seeks to minimize the prediction 
error across all \(N\) instances. We leveraged the information extraction capabilities of large language models (LLMs) for our task. Acknowledging the challenge of adapting these models to domain-specific applications, we employed QLoRA \cite{Qlora}, a recent and parameter-efficient approach that uses 4-bit quantization to enable fine-tuning of large LLMs within existing hardware constraints. QLoRA modules introduce trainable parameters in the form of low-rank matrices. These modules adapt the model for specific tasks without significantly altering the pre-trained weights:
     \(
     \Delta W = A B, 
     \)
     where \(A\) and \(B\) are low-rank matrices. 

\noindent{\textbf{II. Drug Grouping Module: }}
This module groups adverse drug events (ADEs) information based on semantic similarity using a pre-trained language model Sentence-BERT \cite{sentencebert} and categorizing them by severity for each drug. These semantic representations are then clustered using hierarchical clustering to group similar severity side effects. The method organizes each drug's side effects into distinct groups based on severity levels (`High', `Moderate', or `Low'), ensuring specifically grouping of related severity side effects.

\noindent{\textbf{III. ADE Summarization Unit: }}
To efficiently leverage the structured drug information derived from the Drug Grouping Grid, we utilize the T5 Large \cite{flant5} encoder-decoder architecture. The encoder maps the grouped ADE information \( X \) into a fixed-dimensional latent representation, 
   \(
   Z = \text{Encoder}(X). 
   \)
   It employs self-attention and positional encodings to model the intricate relationships between input elements, such as drug names and ADE severity:
   \begin{equation}
   Z_i = \sum_{h=1}^{H} \text{Attention}(X; W^Q_h, W^K_h, W^V_h) + \text{PositionalEncoding}(X)
   \end{equation}
   where \( H \) denotes the number of attention heads, and \( W^Q_h \), \( W^K_h \), \( W^V_h \) are the query, key, and value weight matrices for the \( h \)-th head. The attention mechanism allows the encoder to emphasize crucial input aspects (e.g., notable ADEs and their severity levels):
   \begin{equation}
   \text{Attention}(X; W^Q, W^K, W^V) = \text{softmax}\left(\frac{QK^T}{\sqrt{d_k}}\right)V
   \end{equation}
   where \( Q = XW^Q \), \( K = XW^K \), \( V = XW^V \), and \( d_k \) is the dimension of the key vectors. After the self-attention operation, the output is passed through the Feed Forward Network (FFN) block in the Encoder. This layer further processes the representations to refine the encoded features. The combination of self-attention and FFN operations produces the fixed-size representation \( Z \).
   \begin{equation}  
   Z = \text{SelfAttention}(X) + \text{PositionalEncoding}(X)  
   \end{equation}
   The encoded representation \(Z\) is then passed into the Decoder. As shown in the Figure \ref{gascade}, the Decoder then utilizes cross-attention with \( Z \) to produce the output summary \( Y \).
   It sequentially generates each token \( y_i \) of the output, attending to relevant portions of \( Z \) as it constructs each word:
   \begin{equation}
   y_i = \text{softmax}\left(\sum_{h=1}^{H} \text{CrossAttention}(Z, Y_{<i}; W^Q_h, W^K_h, W^V_h)\right)
   \end{equation}
   where \( Y_{<i} \) represents the sequence of previously generated tokens. The cross-attention mechanism directs the decoder to focus on the most pertinent features of \( Z \) while generating each output word:
   \begin{equation}
   \text{CrossAttention}(Z, Y_{<i}; W^Q, W^K, W^V) = \text{softmax}\left(\frac{QK^T}{\sqrt{d_k}}\right)V
   \end{equation}
   with \( Q = Y_{<i}W^Q \), \( K = ZW^K \), \( V = ZW^V \). The final output \( Y \) is an abstractive summary that presents the ADE information for each drug, prioritizing the highest severity cases first, followed by those with moderate and low severity.

\noindent{\textbf{IV. Alignment using DPO: }}The \textit{GASCADE} model, developed using both the encoder and decoder components of the T5 architecture, employs Direct Preference Optimization (DPO) \cite{DPO} to enhance the quality of its generated summaries. For applying DPO, we created reference summaries crafted by medical interns under the supervision of a medical professional. As non-preferred summaries were not readily available, we generated these using GPT-4o mini \cite{gpt4} model. We then combined the preferred and non-preferred summaries into a synthetic dataset, which was used to train the \textit{GASCADE} model to produce more refined and nuanced summaries.
Direct Preference Optimization (DPO) is based on a loss function where a large language model (LLM) differentiates between a preferred output \( y_w \) and a less preferred output \( y_l \). The DPO loss function is formulated as follows:
\begin{align}
L_{\text{DPO}}(\pi_\theta; \pi_{\text{LLM}}) = - \mathbb{E}_{(x, y_w, y_l) \sim D_{\text{LLM}}} \Bigg[ \log \sigma \Bigg( \beta \log \frac{\pi_\theta(y_w | x)}{\pi_{\text{LLM}}(y_w | x)} \nonumber \\
- \beta \log \frac{\pi_\theta(y_l | x)}{\pi_{\text{LLM}}(y_l | x)} \Bigg) \Bigg]
\end{align}
where:
\begin{itemize}
    \item \( L_{\text{DPO}}(\pi_\theta; \pi_{\text{LLM}}) \) represents the DPO loss function.
    \item \( \mathbb{E}_{(x, y_w, y_l) \sim D_{\text{LLM}}} \) denotes the expectation over the dataset \( D_{\text{LLM}} \), with \( y_w \) as the preferred output and \( y_l \) as the less preferred output, based on the LLM’s preference.
    \item \( \pi_\theta(y_w | x) \) and \( \pi_\theta(y_l | x) \) are the probabilities assigned by the model's policy \( \pi_\theta \) to the preferred and less preferred outputs, respectively, for a given input \( x \).
    \item \( \pi_{\text{LLM}}(y_w | x) \) and \( \pi_{\text{LLM}}(y_l | x) \) represent the probabilities assigned by the LLM to the preferred and less preferred outputs, respectively, for the input \( x \).
    \item \( \beta \) is a scaling factor that controls the sharpness of the preference distinction between the preferred and less preferred outputs.
    \item \( \sigma \) is the sigmoid function, defined as \( \sigma(z) = \frac{1}{1 + e^{-z}} \). It normalizes the difference in log probabilities, yielding a value between 0 and 1 that reflects the likelihood of \( y_w \) being favored over \( y_l \).
\end{itemize}

\section{Experiment Results and Analysis}
This section outlines our experiments, results, and analysis of the proposed framework. The experiments are designed to answer the following research questions: \textit{ RQ1) How does the performance of our proposed framework, \textit{GASCADE}, compare to baseline models? RQ2) What is the effect of employing a two-step process for grouping and abstractive summarization of cancer-related adverse drug events? RQ3) To what degree do alignment algorithms, such as DPO, influence the quality of summaries, especially when preference datasets are generated by Large Language Models?}

\subsection{Experimental Setup}

In the following section, we present our experimental framework and evaluate the effectiveness of the proposed \textit{GASCADE} model using both automated and human-centric assessments. For the summarization step, we employed T5 Large \cite{flant5} as the foundation model, chosen for its ability to understand and align with task-specific instructions, making it well-suited for summarization tasks. Automatic evaluation of the information extraction step includes accuracy, precision, recall, and F1 score for ADE severity prediction. For entity extraction tasks, such as drug names and ADEs, we assess performance using Jaccard Similarity, Hamming Distance (HD), and Ratcliff-Obershelp Similarity (ROS). Summarization quality is measured with ROUGE \cite{lin2004rouge}, BLEU \cite{bleu}, BERTScore \cite{bertscore}, and METEOR \cite{meteor}. Human evaluation employs clinical evaluation score \cite{clinicalevalscore}, factual recall \cite{clinical_metric}, and omission rate \cite{clinical_metric}.
Experiments were conducted using an RTX 6000 GPU, with model runtimes of 30-40 minutes. The dataset was split into 80\% for training, 5\% for validation, and 15\% for testing.



\subsection{\textbf{ Baseline Setup}}
{\textbf{ADE Information Extraction Module}}: We have leveraged the most recent and popular open source LLMs including T5 Large \cite{flant5}, Llama 3 \cite{llama3modelcard} (8b) 16bf, MedLlama \cite{llama3modelcard}, Phi3 (3.8b) \cite{abdin2024phi3technicalreporthighly}, Gemma 2(2b) \cite{gemma_2024} for the information extraction module. These models were fine-tuned on the proposed \textit{MCADRS} dataset, using posts as input sequences and labeled outputs as target sequences, with the training objective defined in Equation 1. 
\newline \textbf{ADE Summarization Unit:} For ADE summarization, we used baseline models such as T5 \cite{flant5}, Gemma \cite{gemma_2024}, Phi3 \cite{abdin2024phi3technicalreporthighly}, Mistral \cite{MISTRAL}, Medalpaca \cite{medalpaca}, and Zephyr \cite{tunstall2023zephyr} on our proposed dataset. In recent years, these models have shown significant success in various summarization tasks. 

\begin{table}[!ht]
\centering
\resizebox{\columnwidth}{!}{%
\begin{tabular}{ccccccccccc}
\hline
\textbf{Models} &  & \textbf{Rouge 1} & \textbf{Rouge 2} & \textbf{Rouge L} & \textbf{Bleu1} & \textbf{Bleu 2} & \textbf{Bleu 3} & \textbf{Bleu 4} & \textbf{Bert Score} & \textbf{Meteor} \\ \hline
\multirow{3}{*}{\textbf{T5 Large    }} & High & \textbf{0.286} & \textbf{0.130} & \textbf{0.260} & \textbf{0.492} & \textbf{0.307} & \textbf{0.203} & \textbf{0.165} & {0.831} & \textbf{0.260} \\
 & Moderate & \textbf{0.327} & \textbf{0.161} & \textbf{0.302} & \textbf{0.517} & \textbf{0.336} & \textbf{0.236} & \textbf{0.199} & 0.840 & \textbf{0.288} \\
 & Mild & 0.166 & 0.029 & \textbf{0.157} & \textbf{0.169} & \textbf{0.106} & \textbf{0.098} & \textbf{0.061} & 0.798 & \textbf{0.095} \\ \hline
\textit{\textbf{}} &  &  &  &  &  &  &  &  &  &  \\
\multirow{3}{*}{\textbf{BART}} & High & 0.246 & 0.119 & 0.202 & 0.302 & 0.209 & 0.147 & 0.122 & 0.837 & 0.150 \\
 & Moderate & 0.275 & 0.143 & 0.236 & 0.365 & 0.247 & 0.171 & 0.141 & 0.843 & 0.154 \\
 & Mild & 0.099 & 0.023 & 0.073 & 0.166 & 0.089 & 0.043 & 0.029 & 0.817 & 0.059 \\ \hline
\textit{\textbf{}} &  &  &  &  &  &  &  &  &  &  \\
\multirow{3}{*}{\textbf{GPT}} & High & 0.276 & 0.109 & 0.180 & 0.104 & 0.097 & 0.081 & 0.068 & \textbf{0.880} & 0.156 \\
 & Moderate & 0.277 & 0.112 & 0.190 & 0.085 & 0.079 & 0.066 & 0.055 & \textbf{0.884} & 0.155 \\
 & Mild & \textbf{0.177} & \textbf{0.057} & 0.149 & 0.009 & 0.008 & 0.007 & 0.006 & \textbf{0.868} & 0.082 \\ \hline
\textit{\textbf{}} &  &  &  &  &  &  &  &  &  &  \\
\multirow{3}{*}{\textbf{Llama 3}} & High & 0.264 & 0.096 & 0.181 & 0.160 & 0.131 & 0.098 & 0.077 & 0.877 & 0.149 \\
 & Moderate & 0.260 & 0.096 & 0.181 & 0.126 & 0.104 & 0.077 & 0.061 & 0.878 & 0.143 \\
 & Mild & 0.172 & 0.053 & 0.145 & 0.105 & 0.072 & 0.045 & 0.033 & 0.865 & 0.091 \\ \hline
 &  &  &  &  &  &  &  &  &  & 
\end{tabular}%
}
\caption{Results of different models on summarisation tasks for different severity.}
\label{summarisation_baseline}
\vspace{-1cm}
\end{table}

\subsection{Automated Evaluation }
Our \textit{GASCADE} framework follows a two-step approach, beginning with ADR information extraction. As shown in Table \ref{Classification}, the base versions of models generally display lower performance, with reduced precision and F1 scores, higher Hamming distances (HD), and lower ROS values. Fine-tuning significantly enhances these metrics, indicating that task-specific adjustments greatly benefit model accuracy. Among the models tested, Phi3 exhibited the strongest performance for ADR information extraction. Following the extraction phase, the framework proceeds to the summarization step. The automatic evaluation results presented in Table \ref{summarisation_baseline} reveal that T5 Large consistently surpasses other models across nearly all metrics and severity levels. Although BART, GPT, and Llama 3 show strong semantic similarity, as indicated by their high BERTScores, they lag in structural alignment, resulting in suboptimal performance in producing contextually relevant summaries.
\begin{table}[ht!]
\tiny
\centering
\resizebox{\columnwidth}{!}{%
\begin{tabular}{cccccclccclccc}
\hline
\multicolumn{1}{l}{} & \multicolumn{1}{l}{} & \multicolumn{4}{c}{\textbf{Severity}} &  & \multicolumn{3}{c}{\textbf{Drug}} &  & \multicolumn{3}{c}{\textbf{ADE}} \\ \hline
\textbf{Models} & \multicolumn{1}{l}{} & A & P & R & F1 & \multicolumn{1}{c}{} & JS & HD & ROS & \multicolumn{1}{c}{} & JS & HD & ROS. \\ \hline
\multirow{2}{*}{\textbf{T5 Large}} & Base & 0.56 & 0.29 & 0.31 & 0.41 &  & 0.23 & 70.38 & 0.38 &  & 0.12 & 65.42 & 0.44 \\
 & FineTuned & 0.79 & 0.71 & 0.69 & 0.64 &  & 0.57 & 5.55 & 0.82 &  & 0.68 & 16.23 & 0.78 \\ \hline
\multicolumn{1}{l}{} & \multicolumn{1}{l}{} &  &  &  &  &  &  &  &  &  &  &  &  \\
\multirow{2}{*}{\textbf{Llama 3}} & Base & 0.59 & 0.30 & 0.34 & 0.28 &  & 0.25 & 94.88 & 0.45 &  & 0.11 & 46.42 & 0.10 \\
 & FineTuned & 0.76 & 0.88 & 0.77 & 0.79 &  & 0.71 & 9.93 & 0.81 &  & 0.65 & 12.37 & 0.79 \\ \hline
\multicolumn{1}{l}{} & \multicolumn{1}{l}{} &  &  &  &  &  &  &  &  &  &  &  &  \\
\multirow{2}{*}{\textbf{MedLlama}} & Base & 0.54 & 0.26 & 0.13 & 0.16 &  & 0.27 & 85.06 & 0.47 &  & 0.19 & 84.57 & 0.39 \\
 & FineTuned & 0.80 & \textbf{0.89} & 0.65 & 0.57 &  & 0.41 & 7.16 & 0.67 &  & 0.30 & 28.36 & 0.50 \\ \hline
\multicolumn{1}{l}{} & \multicolumn{1}{l}{} &  &  &  &  &  &  &  &  &  &  &  &  \\
\multirow{2}{*}{\textbf{Phi3}} & Base & 0.57 & 0.15 & 0.16 & 0.15 &  & 0.08 & 80.46 & 0.17 &  & 0.02 & 30.62 & 0.11 \\
 & FineTuned & \textbf{0.89} & 0.80 & \textbf{0.87} & \textbf{0.86} &  & \textbf{0.96} & \textbf{5.25} & \textbf{0.97} &  & \textbf{0.86} & \textbf{8.27} & \textbf{0.87} \\ \hline
\multicolumn{1}{l}{} & \multicolumn{1}{l}{} &  &  &  &  &  &  &  &  &  &  &  &  \\
\multirow{2}{*}{\textbf{Gemma 2}} & Base & 0.47 & 0.57 & 0.63 & 0.61 &  & 0.04 & 185.27 & 0.27 &  & 0.01 & 179.12 & 0.22 \\
 & FineTuned & 0.83 & 0.82 & 0.78 & 0.69 &  & 0.87 & 7.91 & 0.89 &  & 0.81 & 15.31 & 0.83 \\ \hline
\multicolumn{1}{l}{} & \multicolumn{1}{l}{} &  &  &  &  &  &  &  &  &  &  &  & 
\end{tabular}%
}
\caption{Results of Adverse Drug Information Extraction task for extracting Drug Names, ADE, and Severity Labels. For the ADE and Drug, the results are in terms of JS: Jaccard Similarity, HD: Hamming Distance, and ROS: Ratcliff-Obershelp Similarity. For Severity, the results are in terms of Accuracy (A), Precision (P), Recall (R) and, F1 Scores. }
\label{Classification}
\vspace{-1cm}
\end{table}
\newline{\textbf{RQ1: }}
As can be observed from Table \ref{With_DPO}, the proposed model outperforms all other baselines in the grouped summarization task, demonstrating the advantages of encoder-decoder models in summarization. BART showed the lowest performance, while GPT outperformed the other models. Although Llama 3 performed comparably to GPT, \textit{GASCADE} surpassed all others, establishing itself as the leading performer.
\newline\textbf{RQ2 Ablation Study: }
As shown in Table \ref{without_first_step}, omitting the two-step process that includes ADE information extraction results in a substantial performance decline, with Rouge 1 scores dropping by up to 69.75\% for T5 Large and by 32.06\% for Zephyr. The two-step approach is crucial for generating summaries that capture relevant pharmacovigilance information accurately. Additionally, without this process, LLMs encounter limitations due to context length constraints, despite GPT’s 128K-token capacity, which remains insufficient for large datasets. Therefore, combining ADE extraction with summarization is essential to produce accurate summaries.
\begin{figure}[!htbp]
\centerline{\includegraphics[scale=0.05]{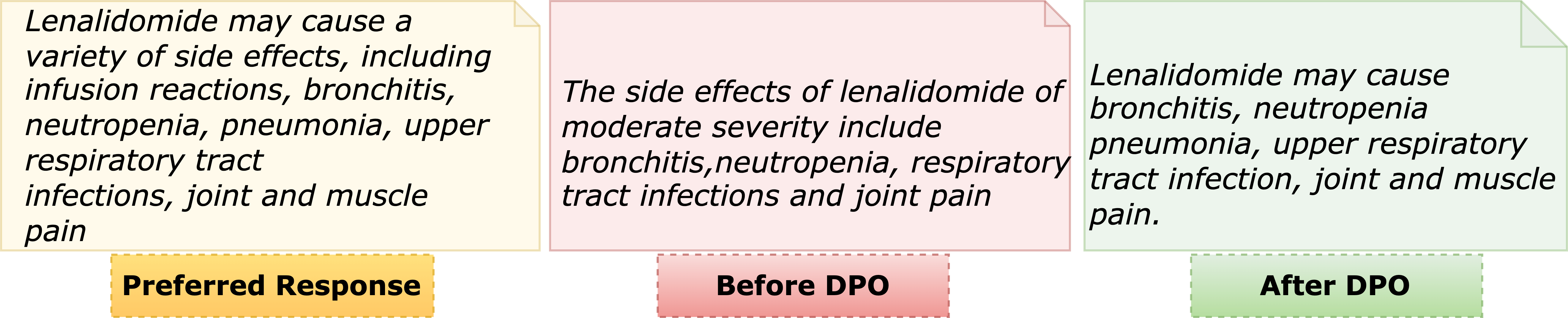}}
\caption{Samples summaries generated with and without alignment techniques}
\label{dpo_comparison}
\end{figure}
\begin{table}[ht!]
\tiny
\centering
\resizebox{\columnwidth}{!}{%
\begin{tabular}{lcccccccc}
\hline
\textbf{Models} & \multicolumn{1}{l}{\textbf{Rouge 1}} & \multicolumn{1}{l}{\textbf{Rouge 2}} & \multicolumn{1}{l}{\textbf{Rouge L}} & \multicolumn{1}{l}{\textbf{Bleu 1}} & \multicolumn{1}{l}{\textbf{Bleu 2}} & \multicolumn{1}{l}{\textbf{Bleu 3}} & \multicolumn{1}{l}{\textbf{Bert Score}} & \multicolumn{1}{l}{\textbf{Meteor}} \\ \hline
\textbf{T5 Large} & 0.0900 & 0.0105 & 0.0738 & 0.0464 & 0.0094 & 0.0027 & 0.1251 & 0.0799 \\
\textbf{Phi3} & 0.1512 & 0.0267 & 0.1333 & 0.0925 & 0.0329 & 0.0093 & 0.2393 & 0.1608 \\
\textbf{Gemma} & 0.0677 & 0.0096 & 0.0578 & 0.0308 & 0.0095 & 0.0019 & 0.0867 & 0.0618 \\
\textbf{Mistral} & 0.1188 & 0.0204 & 0.1044 & 0.0631 & 0.0209 & 0.0048 & 0.1564 & 0.1081 \\
\textbf{Medalpaca} & 0.1217 & 0.0215 & 0.1027 & 0.0657 & 0.0207 & 0.0059 & 0.2143 & 0.1323 \\
\textbf{Zephyr} & \textbf{0.1690} & \textbf{0.0373} & \textbf{0.1435} & \textbf{0.1074} & \textbf{0.0434} & \textbf{0.0155} & \textbf{0.2535} & \textbf{0.1833} \\ \hline
\end{tabular}%
}
\vspace{-0.06cm}
\caption{Results of different models without information extraction. (Ablation Study)}
\label{without_first_step}
\vspace{-0.5cm}
\end{table}
\newline \textbf{RQ3: }As demonstrated in Table \ref{With_DPO} and Figure \ref{dpo_comparison}, our tests reveal substantial performance improvements with the implementation of DPO. This performance gain can be attributed to DPO's direct alignment with user-defined objectives, minimizing irrelevant or repetitive content, thereby producing preferred summaries.


\begin{table}[ht!]
\centering
\resizebox{\columnwidth}{!}{%
\begin{tabular}{c|c|cccccccc}
\hline
\multicolumn{1}{l|}{\textbf{}} & \multicolumn{1}{l|}{\textbf{}} & \multicolumn{1}{l}{\textbf{Rouge 1}} & \multicolumn{1}{l}{\textbf{Rouge 2}} & \multicolumn{1}{l}{\textbf{Rouge L}} & \multicolumn{1}{l}{\textbf{Bleu 1}} & \multicolumn{1}{l}{\textbf{Bleu 2}} & \multicolumn{1}{l}{\textbf{Bleu 3}} & \multicolumn{1}{l}{\textbf{Bert Score}} & \multicolumn{1}{l}{\textbf{Meteor}} \\ \hline
\multirow{2}{*}{\textbf{GASCADE (Ours)}} & \textbf{Base} & 0.2971 & 0.1405 & 0.2721 & 0.1997 & 0.1281 & 0.0667 & 0.3048 & 0.2676 \\
 & \textbf{After DPO} & \textbf{0.3926} & \textbf{0.1798} & \textbf{0.3501} & \textbf{0.2574} & \textbf{0.1650} & \textbf{0.0986} & \textbf{0.4171} & 0.3678 \\ \cline{1-1}
\multicolumn{1}{l|}{} & \multicolumn{1}{l|}{} & \multicolumn{1}{l}{} & \multicolumn{1}{l}{} & \multicolumn{1}{l}{} & \multicolumn{1}{l}{} & \multicolumn{1}{l}{} & \multicolumn{1}{l}{} & \multicolumn{1}{l}{} & \multicolumn{1}{l}{} \\
\multirow{2}{*}{\textbf{Gemma}} & \textbf{Base} & 0.1669 & 0.0553 & 0.1311 & 0.0942 & 0.0514 & 0.0234 & 0.2606 & 0.2852 \\
 & \textbf{After DPO} & 0.2476 & 0.1025 & 0.2049 & 0.1389 & 0.0858 & 0.0471 & 0.3278 & 0.3930 \\ \cline{1-1}
\multicolumn{1}{l|}{} & \multicolumn{1}{l|}{} & \multicolumn{1}{l}{} & \multicolumn{1}{l}{} & \multicolumn{1}{l}{} & \multicolumn{1}{l}{} & \multicolumn{1}{l}{} & \multicolumn{1}{l}{} & \multicolumn{1}{l}{} & \multicolumn{1}{l}{} \\
\multirow{2}{*}{\textbf{Phi3}} & \textbf{Base} & 0.1362 & 0.0401 & 0.1082 & 0.0720 & 0.0366 & 0.0163 & 0.2481 & 0.2801 \\
 & \textbf{After DPO} & 0.2188 & 0.0963 & 0.1860 & 0.1198 & 0.0770 & 0.0467 & 0.3203 & 0.3779 \\ \cline{1-1}
\multicolumn{1}{l|}{} & \multicolumn{1}{l|}{} & \multicolumn{1}{l}{} & \multicolumn{1}{l}{} & \multicolumn{1}{l}{} & \multicolumn{1}{l}{} & \multicolumn{1}{l}{} & \multicolumn{1}{l}{} & \multicolumn{1}{l}{} & \multicolumn{1}{l}{} \\
\multirow{2}{*}{\textbf{Mistral}} & \textbf{Base} & 0.1498 & 0.0505 & 0.1216 & 0.0841 & 0.0459 & 0.0224 & 0.2559 & 0.2780 \\
 & \textbf{After DPO} & 0.2684 & 0.1184 & 0.2262 & 0.1514 & 0.0968 & 0.0569 & 0.3387 & \textbf{0.3982} \\ \cline{1-1}
\multicolumn{1}{l|}{} & \multicolumn{1}{l|}{} & \multicolumn{1}{l}{} & \multicolumn{1}{l}{} & \multicolumn{1}{l}{} & \multicolumn{1}{l}{} & \multicolumn{1}{l}{} & \multicolumn{1}{l}{} & \multicolumn{1}{l}{} & \multicolumn{1}{l}{} \\
\multirow{2}{*}{\textbf{Medalpaca}} & \textbf{Base} & 0.1230 & 0.0370 & 0.0999 & 0.0653 & 0.0338 & 0.0157 & 0.2317 & 0.2491 \\
 & \textbf{After DPO} & 0.1915 & 0.0639 & 0.1458 & 0.0985 & 0.0520 & 0.0280 & 0.2996 & 0.3003 \\ \cline{1-1}
\multicolumn{1}{l|}{} & \multicolumn{1}{l|}{} & \multicolumn{1}{l}{} & \multicolumn{1}{l}{} & \multicolumn{1}{l}{} & \multicolumn{1}{l}{} & \multicolumn{1}{l}{} & \multicolumn{1}{l}{} & \multicolumn{1}{l}{} & \multicolumn{1}{l}{} \\
\multirow{2}{*}{\textbf{Zephyr}} & \textbf{Base} & 0.1697 & 0.0568 & 0.1380 & 0.0910 & 0.0501 & 0.0240 & 0.2739 & 0.2645 \\
 & \multicolumn{1}{l|}{\textbf{After DPO}} & 0.2498 & 0.1053 & 0.2091 & 0.1371 & 0.0866 & 0.0486 & 0.3583 & 0.3694 \\ \hline
\multicolumn{1}{l}{} & \multicolumn{1}{l}{} & \multicolumn{1}{l}{} & \multicolumn{1}{l}{} & \multicolumn{1}{l}{} & \multicolumn{1}{l}{} & \multicolumn{1}{l}{} & \multicolumn{1}{l}{} & \multicolumn{1}{l}{} & \multicolumn{1}{l}{} \\ 
\end{tabular}%
}
\caption{Results of different baselines on the grouped summarisation task with DPO alignment.}
\label{With_DPO}
\vspace{-0.5cm}
\end{table}


\subsection{Human Evaluation}
Automatic evaluation metrics often fall short in the medical domain, where inaccuracies or omissions can have severe consequences. To address this, a team of three medical experts, conducted a human evaluation on randomly selected 20\% of the test samples. Evaluation criteria included the Clinical Evaluation Score \cite{clinicalevalscore}, where doctors and team members rated summaries from 1 (poor) to 5 (excellent) based on relevance, consistency, fluency, coherence, and degree of hallucination. Additionally, Medical Fact-Based Metrics, such as Factual Recall \cite{clinical_metric} and Omission Rate \cite{clinical_metric}, assessed the accuracy of generated summaries in capturing essential medical facts compared to the gold-standard annotated summary. Table \ref{expert_scores_image} presents the results for \textit{GASCADE}, highlighting its significant outperformance of all baselines across the chosen human evaluation metrics. 



\begin{table}[ht!]
\tiny
\centering
\resizebox{\textwidth}{!}{%
\begin{tabular}{cccc}
\hline
\multicolumn{1}{l}{} & \multicolumn{1}{l}{\textbf{Clinical Evaluation Score}} & \multicolumn{1}{l}{\textbf{Factual Recall}} & \multicolumn{1}{l}{\textbf{Omission Rate}} \\ \hline
GASCADE (Ours) & 3.88 & 0.85 & 0.15 \\
Gemma & 2.91 & 0.61 & 0.33 \\
Phi3 & 3.13 & 0.69 & 0.21 \\
Mistral & 3.37 & 0.78 & 0.19 \\
Medalpaca & 2.82 & 0.65 & 0.35 \\
Zephyr & 3.34 & 0.73 & 0.2 \\
\textbf{Annotated Summary} & \textbf{4.21} & \textbf{0.89} & \textbf{0.09} \\ \hline
\end{tabular}%
}
\vspace{0.1mm}
\caption{Human evaluation scores of various models across different metrics.}
\label{expert_scores_image}
\vspace{-0.9cm}
\end{table}

\subsection{Qualitative evaluation }
\begin{figure}[!htbp]
\vspace{-0.2cm}
\centerline{\includegraphics[scale=0.04]{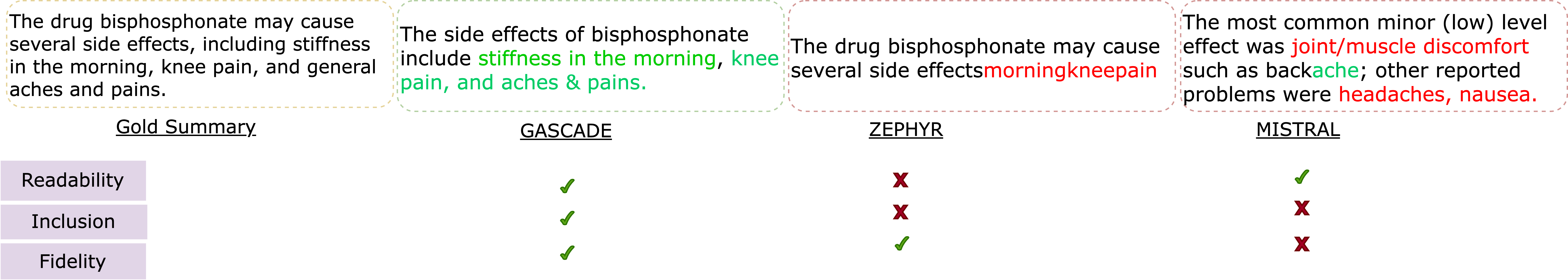}}
\caption{Samples responses generated via our framework and other models}
\label{qualitative_comparison}
\vspace{-1cm}
\end{figure}

Figure \ref{qualitative_comparison} presents a comparative analysis of summaries produced by the top three models in terms of performance. This evaluation highlights three key dimensions: (1) Readability, assessing whether the summaries are clear, concise, and suitable for clinical interpretation; (2) Inclusion, determining if all relevant ADEs are captured within the summary; and (3) Fidelity, ensuring the summary accurately reflects the original information without inaccuracies or omissions. \textit{GASCADE} demonstrates strong performance, producing summaries that are highly readable, inclusive, and faithful to the source. In contrast, Zephyr often lacks clarity and omits important ADEs, while Mistral tends to introduce extra information and exhibits some hallucination.

\section{Ethical Consideration}

\begin{itemize}
   \item {\textbf{User Privacy}}
Our dataset includes ADE, drug names, and corresponding posts, each annotated with labels, while ensuring no personal user information is included.
\item {\textbf{Biases}}
Any biases identified within the dataset are unintentional. We emphasize our commitment to not causing harm to any individual or group. Recognizing the subjectivity involved in assessing whether a post contains ADE, we have obtained consensus from all annotators before finalizing the data.
\item {\textbf{Intended Use}}
We provide this dataset to encourage further research in the detection of Adverse Drug Events. It is shared exclusively for research purposes and is not licensed for commercial use.

\end{itemize}

\section{Conclusion and Future work}

This study introduces a novel task of Grouped ADE Summarization for pharmacovigilance, marking the first attempt at summarizing ADRs in the field of cancer care. Our approach features a unique two-step framework, combining an ADR information extraction module with a summarization module. We also present the \textit{MCADRS} dataset, the first dataset specifically focusing on ADRs associated with cancer drugs. Our pipeline architecture, \textit{GASCADE}, demonstrates superior performance across all baselines according to both human and automated evaluation metrics. These findings underscore the effectiveness of alignment techniques like DPO in enhancing summarization quality, leading to more refined and nuanced summaries.

\section{Acknowledgements}

Sofia Jamil acknowledges the prestigious Prime
Minister’s Research Fellowship (PMRF) by the
Government of India for providing financial support during this research. We also thank our colleagues at the Department of Computer Science
and Engineering, Indian Institute of Technology
Patna, for their continuous support and valuable
feedback.


%
%
%
%

\bibliography{tai_ref}

\begin{thebibliography}{10}

\bibitem{clinical_metric}
A.~B. Abacha, W.~wai Yim, G.~Michalopoulos, and T.~Lin.
\newblock An investigation of evaluation metrics for automated medical note generation, 2023.

\bibitem{abdin2024phi3technicalreporthighly}
M.~Abdin, J.~Aneja, H.~Awadalla, A.~Awadallah, A.~A. Awan, N.~Bach, A.~Bahree, A.~Bakhtiari, J.~Bao, H.~Behl, A.~Benhaim, M.~Bilenko, J.~Bjorck, S.~Bubeck, M.~Cai, Q.~Cai, V.~Chaudhary, D.~Chen, D.~Chen, W.~Chen, Y.-C. Chen, Y.-L. Chen, H.~Cheng, P.~Chopra, X.~Dai, M.~Dixon, R.~Eldan, V.~Fragoso, J.~Gao, M.~Gao, M.~Gao, A.~Garg, A.~D. Giorno, A.~Goswami, S.~Gunasekar, E.~Haider, J.~Hao, R.~J. Hewett, W.~Hu, J.~Huynh, D.~Iter, S.~A. Jacobs, M.~Javaheripi, X.~Jin, N.~Karampatziakis, P.~Kauffmann, M.~Khademi, D.~Kim, Y.~J. Kim, L.~Kurilenko, J.~R. Lee, Y.~T. Lee, Y.~Li, Y.~Li, C.~Liang, L.~Liden, X.~Lin, Z.~Lin, C.~Liu, L.~Liu, M.~Liu, W.~Liu, X.~Liu, C.~Luo, P.~Madan, A.~Mahmoudzadeh, D.~Majercak, M.~Mazzola, C.~C.~T. Mendes, A.~Mitra, H.~Modi, A.~Nguyen, B.~Norick, B.~Patra, D.~Perez-Becker, T.~Portet, R.~Pryzant, H.~Qin, M.~Radmilac, L.~Ren, G.~de~Rosa, C.~Rosset, S.~Roy, O.~Ruwase, O.~Saarikivi, A.~Saied, A.~Salim, M.~Santacroce, S.~Shah, N.~Shang, H.~Sharma, Y.~Shen, S.~Shukla, X.~Song, M.~Tanaka,
  A.~Tupini, P.~Vaddamanu, C.~Wang, G.~Wang, L.~Wang, S.~Wang, X.~Wang, Y.~Wang, R.~Ward, W.~Wen, P.~Witte, H.~Wu, X.~Wu, M.~Wyatt, B.~Xiao, C.~Xu, J.~Xu, W.~Xu, J.~Xue, S.~Yadav, F.~Yang, J.~Yang, Y.~Yang, Z.~Yang, D.~Yu, L.~Yuan, C.~Zhang, C.~Zhang, J.~Zhang, L.~L. Zhang, Y.~Zhang, Y.~Zhang, Y.~Zhang, and X.~Zhou.
\newblock Phi-3 technical report: A highly capable language model locally on your phone, 2024.

\bibitem{gpt4}
J.~Achiam, S.~Adler, S.~Agarwal, L.~Ahmad, I.~Akkaya, F.~L. Aleman, D.~Almeida, J.~Altenschmidt, S.~Altman, S.~Anadkat, et~al.
\newblock Gpt-4 technical report.
\newblock {\em arXiv preprint arXiv:2303.08774}, 2023.

\bibitem{llama3modelcard}
AI@Meta.
\newblock Llama 3 model card.
\newblock 2024.

\bibitem{meteor}
S.~Banerjee and A.~Lavie.
\newblock Meteor: An automatic metric for mt evaluation with improved correlation with human judgments.
\newblock In {\em Proceedings of the acl workshop on intrinsic and extrinsic evaluation measures for machine translation and/or summarization}, pages 65--72, 2005.

\bibitem{rl21}
R.~Daniulaityte, L.~Chen, F.~R. Lamy, R.~G. Carlson, K.~Thirunarayan, A.~Sheth, et~al.
\newblock “when ‘bad’is ‘good’”: identifying personal communication and sentiment in drug-related tweets.
\newblock {\em JMIR public health and surveillance}, 2(2):e6327, 2016.

\bibitem{Qlora}
T.~Dettmers, A.~Pagnoni, A.~Holtzman, and L.~Zettlemoyer.
\newblock Qlora: Efficient finetuning of quantized llms.
\newblock {\em Advances in Neural Information Processing Systems}, 36, 2024.

\bibitem{cdc}
C.~for Disease~Control, Prevention, et~al.
\newblock Adverse drug events in adults.
\newblock {\em CDC. Retrieved November}, 19:2022, 2017.

\bibitem{ghosh_question}
A.~Ghosh, A.~Acharya, R.~Jain, S.~Saha, A.~Chadha, and S.~Sinha.
\newblock Clipsyntel: clip and llm synergy for multimodal question summarization in healthcare.
\newblock In {\em Proceedings of the AAAI Conference on Artificial Intelligence}, volume~38, pages 22031--22039, 2024.

\bibitem{ghosh_code_mix}
A.~Ghosh, A.~Acharya, P.~Jha, S.~Saha, A.~Gaudgaul, R.~Majumdar, A.~Chadha, R.~Jain, S.~Sinha, and S.~Agarwal.
\newblock Medsumm: A multimodal approach to summarizing code-mixed hindi-english clinical queries.
\newblock In {\em European Conference on Information Retrieval}, pages 106--120. Springer, 2024.

\bibitem{ghosh_summarisation}
A.~Ghosh, M.~Tomar, A.~Tiwari, S.~Saha, J.~Salve, and S.~Sinha.
\newblock From sights to insights: Towards summarization of multimodal clinical documents.
\newblock In {\em Proceedings of the 62nd Annual Meeting of the Association for Computational Linguistics (Volume 1: Long Papers)}, pages 13117--13129, 2024.

\bibitem{gurulingappa}
H.~Gurulingappa, A.~M. Rajput, A.~Roberts, J.~Fluck, M.~Hofmann-Apitius, and L.~Toldo.
\newblock Development of a benchmark corpus to support the automatic extraction of drug-related adverse effects from medical case reports.
\newblock {\em Journal of biomedical informatics}, 45(5):885--892, 2012.

\bibitem{medalpaca}
T.~Han, L.~C. Adams, J.-M. Papaioannou, P.~Grundmann, T.~Oberhauser, A.~L{\"o}ser, D.~Truhn, and K.~K. Bressem.
\newblock Medalpaca--an open-source collection of medical conversational ai models and training data.
\newblock {\em arXiv preprint arXiv:2304.08247}, 2023.

\bibitem{hou2016national}
Y.~Hou, X.~Li, G.~Wu, and X.~Ye.
\newblock National adr monitoring system in china.
\newblock {\em Drug safety}, 39:1043--1051, 2016.

\bibitem{rl25}
S.~Hussain, H.~Afzal, R.~Saeed, N.~Iltaf, and M.~Y. Umair.
\newblock Pharmacovigilance with transformers: A framework to detect adverse drug reactions using bert fine-tuned with farm.
\newblock {\em Computational and Mathematical Methods in Medicine}, 2021(1):5589829, 2021.

\bibitem{MISTRAL}
A.~Q. Jiang, A.~Sablayrolles, A.~Mensch, C.~Bamford, D.~S. Chaplot, D.~de~las Casas, F.~Bressand, G.~Lengyel, G.~Lample, L.~Saulnier, L.~R. Lavaud, M.-A. Lachaux, P.~Stock, T.~L. Scao, T.~Lavril, T.~Wang, T.~Lacroix, and W.~E. Sayed.
\newblock Mistral 7b, 2023.

\bibitem{rl22}
M.~Joshi, D.~Chen, Y.~Liu, D.~S. Weld, L.~Zettlemoyer, and O.~Levy.
\newblock Spanbert: Improving pre-training by representing and predicting spans.
\newblock {\em Transactions of the association for computational linguistics}, 8:64--77, 2020.

\bibitem{karimi2015text}
S.~Karimi, C.~Wang, A.~Metke-Jimenez, R.~Gaire, and C.~Paris.
\newblock Text and data mining techniques in adverse drug reaction detection.
\newblock {\em ACM Computing Surveys (CSUR)}, 47(4):1--39, 2015.

\bibitem{li}
J.~Li, Y.~Sun, R.~J. Johnson, D.~Sciaky, C.-H. Wei, R.~Leaman, A.~P. Davis, C.~J. Mattingly, T.~C. Wiegers, and Z.~Lu.
\newblock Biocreative v cdr task corpus: a resource for chemical disease relation extraction.
\newblock {\em Database}, 2016, 2016.

\bibitem{rl26}
Z.~Li, Z.~Yang, L.~Luo, Y.~Xiang, and H.~Lin.
\newblock Exploiting adversarial transfer learning for adverse drug reaction detection from texts.
\newblock {\em Journal of biomedical informatics}, 106:103431, 2020.

\bibitem{lin2004rouge}
C.-Y. Lin.
\newblock Rouge: A package for automatic evaluation of summaries.
\newblock In {\em Text summarization branches out}, pages 74--81, 2004.

\bibitem{liu2019towards}
F.~Liu, A.~Jagannatha, and H.~Yu.
\newblock Towards drug safety surveillance and pharmacovigilance: current progress in detecting medication and adverse drug events from electronic health records.
\newblock {\em Drug safety}, 42(1):95--97, 2019.

\bibitem{rl24}
Y.~Mohammadi, F.~Ghasemian, J.~Varshosaz, and M.~Sattari.
\newblock Classifying referring/non-referring adr in biomedical text using deep learning.
\newblock {\em Informatics in Medicine Unlocked}, 39:101246, 2023.

\bibitem{rl20}
A.~Nikfarjam, A.~Sarker, K.~O’connor, R.~Ginn, and G.~Gonzalez.
\newblock Pharmacovigilance from social media: mining adverse drug reaction mentions using sequence labeling with word embedding cluster features.
\newblock {\em Journal of the American Medical Informatics Association}, 22(3):671--681, 2015.

\bibitem{oronoz}
M.~Oronoz, K.~Gojenola, A.~P{\'e}rez, A.~D. De~Ilarraza, and A.~Casillas.
\newblock On the creation of a clinical gold standard corpus in spanish: Mining adverse drug reactions.
\newblock {\em Journal of biomedical informatics}, 56:318--332, 2015.

\bibitem{bleu}
K.~Papineni, S.~Roukos, T.~Ward, and W.-J. Zhu.
\newblock Bleu: a method for automatic evaluation of machine translation.
\newblock In {\em Proceedings of the 40th annual meeting of the Association for Computational Linguistics}, pages 311--318, 2002.

\bibitem{patki}
A.~Patki, A.~Sarker, P.~Pimpalkhute, A.~Nikfarjam, R.~Ginn, K.~O’Connor, K.~Smith, and G.~Gonzalez.
\newblock Mining adverse drug reaction signals from social media: going beyond extraction.
\newblock {\em Proceedings of BioLinkSig}, 2014:1--8, 2014.

\bibitem{DPO}
R.~Rafailov, A.~Sharma, E.~Mitchell, C.~D. Manning, S.~Ermon, and C.~Finn.
\newblock Direct preference optimization: Your language model is secretly a reward model.
\newblock {\em Advances in Neural Information Processing Systems}, 36, 2024.

\bibitem{flant5}
C.~Raffel, N.~Shazeer, A.~Roberts, K.~Lee, S.~Narang, M.~Matena, Y.~Zhou, W.~Li, and P.~J. Liu.
\newblock Exploring the limits of transfer learning with a unified text-to-text transformer.
\newblock {\em Journal of machine learning research}, 21(140):1--67, 2020.

\bibitem{sentencebert}
N.~Reimers.
\newblock Sentence-bert: Sentence embeddings using siamese bert-networks.
\newblock {\em arXiv preprint arXiv:1908.10084}, 2019.

\bibitem{clinicalevalscore}
P.~Sahoo, P.~Meharia, A.~Ghosh, S.~Saha, V.~Jain, and A.~Chadha.
\newblock Unveiling hallucination in text, image, video, and audio foundation models: A comprehensive survey.
\newblock {\em arXiv preprint arXiv:2405.09589}, 2024.

\bibitem{sarker2015portable}
A.~Sarker and G.~Gonzalez.
\newblock Portable automatic text classification for adverse drug reaction detection via multi-corpus training.
\newblock {\em Journal of biomedical informatics}, 53:196--207, 2015.

\bibitem{rl19}
A.~Sarker and G.~Gonzalez.
\newblock Portable automatic text classification for adverse drug reaction detection via multi-corpus training.
\newblock {\em Journal of biomedical informatics}, 53:196--207, 2015.

\bibitem{shareef2015development}
S.~Shareef, C.~Naidu, S.~R. Raikar, Y.~V. Rao, and U.~Devika.
\newblock Development, implementation, and analysis of adverse drug reaction monitoring system in a rural tertiary care teaching hospital in narketpally, telangana.
\newblock {\em Int J Basic Clin Pharmacol}, 4(4):757--60, 2015.

\bibitem{singh2017adverse}
P.~Singh, M.~Agrawal, R.~Hishikar, U.~Joshi, B.~Maheshwari, and A.~Halwai.
\newblock Adverse drug reactions at adverse drug reaction monitoring center in raipur: Analysis of spontaneous reports during 1 year.
\newblock {\em Indian journal of pharmacology}, 49(6):432--437, 2017.

\bibitem{gemma_2024}
G.~Team.
\newblock Gemma.
\newblock 2024.

\bibitem{tunstall2023zephyr}
L.~Tunstall, E.~Beeching, N.~Lambert, N.~Rajani, K.~Rasul, Y.~Belkada, S.~Huang, L.~von Werra, C.~Fourrier, N.~Habib, N.~Sarrazin, O.~Sanseviero, A.~M. Rush, and T.~Wolf.
\newblock Zephyr: Direct distillation of lm alignment, 2023.

\bibitem{kappa}
A.~J. Viera, J.~M. Garrett, et~al.
\newblock Understanding interobserver agreement: the kappa statistic.
\newblock {\em Fam med}, 37(5):360--363, 2005.

\bibitem{bertscore}
T.~Zhang, V.~Kishore, F.~Wu, K.~Q. Weinberger, and Y.~Artzi.
\newblock Bertscore: Evaluating text generation with bert.
\newblock {\em arXiv preprint arXiv:1904.09675}, 2019.

\end{thebibliography}
\bibliographystyle{abbrv}








\end{document}